\documentclass[conference]{IEEEtran}
\IEEEoverridecommandlockouts

\usepackage{cite}
\usepackage{amsmath,amssymb,amsfonts}
\usepackage{algorithmic}
\usepackage{graphicx}
\usepackage{textcomp}
\usepackage{xcolor}

\usepackage{graphics} 
\usepackage{epsfig} 
\usepackage{mathptmx} 
\usepackage{times} 
\usepackage{amsmath} 
\usepackage{amssymb}  
\usepackage{subcaption}
\usepackage{float}

\def\BibTeX{{\rm B\kern-.05em{\sc i\kern-.025em b}\kern-.08em
    T\kern-.1667em\lower.7ex\hbox{E}\kern-.125emX}}
\begin{document}

\title{\LARGE \bf
A Target-Based Extrinsic Calibration Framework for Non-Overlapping Camera-Lidar Systems Using a Motion Capture System
}

\author{
\centering
\IEEEauthorblockN{Nicholas Charron}
\IEEEauthorblockA{\textit{Department of Mechanical and Mechatronics Engineering} \\
\textit{University of Waterloo}\\
Waterloo, Canada \\
ncharron@uwaterloo.ca}
\and
\IEEEauthorblockN{Huaiyuan Weng}
\IEEEauthorblockA{\textit{Department of Civil and Environmental Engineering} \\
\textit{University of Waterloo}\\
Waterloo, Canada \\
huaiyuan.weng@uwaterloo.ca}
\and
\IEEEauthorblockN{Steven L. Waslander}
\IEEEauthorblockA{\textit{University of Toronto Institute for Aerospace Studies} \\
\textit{University of Toronto}\\
Toronto, Canada \\
stevenw@utias.utoronto.ca}
\and
\IEEEauthorblockN{Sriram Narasimhan}
\IEEEauthorblockA{\textit{Department Civil and Environmental Engineering} \\
\textit{University of California Los Angeles}\\
Los Angeles, USA \\
snarasim@ucla.edu}

}

\maketitle

\begin{abstract}
   We present a novel target-based lidar-camera extrinsic calibration methodology that can be used for non-overlapping field of view (FOV) sensors. Contrary to previous work, our methodology overcomes the non-overlapping FOV challenge using a motion capture system (MCS) instead of traditional simultaneous localization and mapping approaches. Due to the high relative precision of MCSs, our methodology can achieve both the high accuracy and repeatable calibrations common to traditional target-based methods, regardless of the amount of overlap in the sensors' field of view. Furthermore, we design a target-agnostic implementation that does not require uniquely identifiable features by using an iterative closest point approach, enabled by the MSC measurements. We show using simulation that we can accurately recover extrinsic calibrations for a range of perturbations to the true calibration that would be expected in real circumstances. We prove experimentally that our method out-performs state-of-the-art lidar-camera extrinsic calibration methods that can be used for non-overlapping FOV systems, while using a target-based approach that guarantees repeatably high accuracy. Lastly, we show in simulation that different target designs can be used, including easily constructed 3D targets such as a cylinder that are normally considered degenerate in most calibration formulations.
\end{abstract}

\begin{IEEEkeywords}
Calibration, 3D Point Clouds, Sensor Fusion.
\end{IEEEkeywords}

\section{Introduction}
\label{sec:intro}

Robotic sensors such as lidars and cameras carry out numerous tasks in modern robotic systems, such as perception, localization and mapping. While these sensors can occasionally be used in isolation, many tasks they support require data from the different sensors to be fused, requiring rigid body transformations (i.e., rotations and translations) between sensor frames to be estimated. Such extrinsic calibration enables 3D data from range sensors to be projected into the 2D image coordinate systems, as well as 2D image data to be back-projected and combined with the 3D sensor data (e.g., point cloud colorization). Obtaining accurate extrinsic calibration is essential to sensor data fusion and to ensure the success of downstream data processing tasks.

In general, extrinsic calibration can be divided into two approaches: target-based and targetless. Target-based calibration methods (~\cite{Zhang2004}, ~\cite{Kim2019}, ~\cite{Geiger2012}, ~\cite{Alismail2012}, ~\cite{RodriguezF.2008}, ~\cite{VelasM2014}, ~\cite{Guindel2018}, ~\cite{Domhof2019}, ~\cite{Kummerle2018}) are performed in a lab setting and rely on extracting distinct features from specially designed targets that are observed by multiple sensors. These methodologies can achieve high accuracy but require an overlap in the field of view (FOV) of the sensors in order to detect the same target from at least two sensors simultaneously. Since sensor overlap often only occurs between small subsets of sensors in larger systems, these methods require calibration transformations to be estimated individually and a chain of transformations between sensors to be generated. This approach can lead to compounding of calibration errors along the chain. The accuracy of the calibration also depends on the amount of overlap between sensors since the measurement sample space is restricted to lie within the overlapping region~\cite{Rebello2020}.

Targetless calibration approaches (~\cite{Castorena2016}, ~\cite{Levinson2013}, ~\cite{Pandey2012}, ~\cite{Taylor2013}, ~\cite{Scott2015}, ~\cite{Napier2013}, ~\cite{Taylor2016}, ~\cite{Andreff2001}, ~\cite{Heller2011}, ~\cite{Jeong2019}) can overcome the sensor overlap requirement by moving the sensors through an environment and performing simultaneous localization and mapping (SLAM). These methods can associate a variety of features in the scene between sensors, construct a map with multiple sensor streams while optimizing over calibration parameters directly, or perform a hand-eye calibration that aligns SLAM trajectories created independently with each sensor. Contrary to target-based calibration methods, the accuracy of these methodologies cannot be guaranteed since corresponding features, building maps without calibration information, or aligning trajectories derived from the lidar and camera data are all susceptible to incorrect correspondence in natural scenes (i.e., without targets), leading to both reduced calibration accuracy and repeatability. 

\begin{figure}[thpb]
  \centering
  \includegraphics[width=0.5\textwidth]{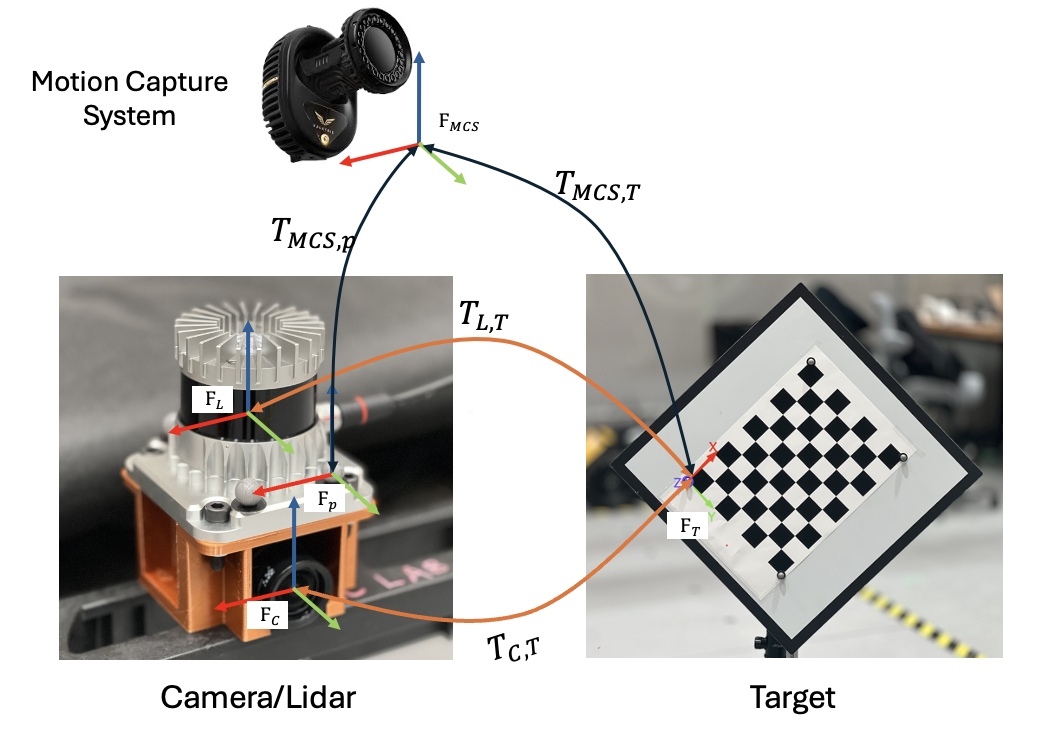}
  \caption{Illustration of the proposed calibration framework with Motion Capture System}
  \label{fig:idea_diagram}
\end{figure}

In this work, we present a novel target-based calibration method that can perform extrinsic calibration of non-overlapping lidars and cameras without the requirement of performing SLAM. As shwon in Fig.~\ref{fig:idea_diagram}, our method uses a motion capture system (MCS)~\cite{Merriaux2017} to track the pose of the sensors and the targets. We can achieve the repeatable high precision benefits of target-based methods, as well as the non-overlapping benefit of the targetless calibration methods by replacing SLAM with a MCS. Since the MCS accuracy is at least an order of magnitude greater than the accuracy of the sensor feature measurements \cite{Merriaux2017}, the additional error introduced by the MCS to overcome the sensor overlap limitation is negligible. 

Our MCS-based methodology also introduces other benefits beyond high accuracy target-based calibrations for non-overlapping cameras. Firstly, since all sensors can be calibrated with respect to one common frame of reference, the accuracy of the calibration is independent of the number of sensors and therefore eliminates the problem of compounding errors from estimating a chain of transformations. Second, since the location of the targets can be estimated relative to the sensors at any time, we can reduce the search space for the targets in the sensor data to eliminate outlier measurements and further increase the calibration accuracy and repeatability of our target-based calibration method. 

To further increase the utility of this approach, we also provide an implementation that is agnostic to target design, or feature extraction methods. We do this by using point-to-point error terms, combining all target observations into a single calibration problem for each sensor, and use an iterative correspondence approach that is only possible with the MCS measurements. This allows any target to be used as long as that target can be extracted from the lidar and camera data, even if this target presents a normally degenerate case that could not fully constrain the calibrations with a single measurement. Despite the large variety of target designs used in previous calibration work, to the best of our knowledge, there exists no mathematical framework or open source software that allows the user to easily implement any target design or feature extraction method of their choice into a calibration framework. 

The specific contributions of this work are as follows:
\begin{itemize}[noitemsep, topsep=0pt]
    \item We present the first target-based calibration method for non-overlapping lidar-camera systems using a MCS that outperforms state-of-the-art methods in accuracy and repeatability.
    \item We create a target-agnostic calibration formulation allowing for various target designs with and without unique feature points.
    \item We provide an extensible open source software\footnote{https://github.com/nickcharron/vicon$\_$calibration} specifically designed for generic lidar \& camera extrinsic calibration with a MCS.
\end{itemize}

Despite the high costs of a typical MCS, the proposed approach has many suitable applications. Firstly, the majority of research labs that work in robotics and computer vision already have a MCS for getting ground truth results for their research projects. Furthermore, many industries have significant budgets for setting up proper calibration labs for performing mass scale calibration such as fleet calibration for automotive industries, or calibration of augmented/virtual reality consumer devices. The combination of increased accuracy and sensor configuration flexibility enabled by the use of the MCS in calibration makes this a promising approach.

\section{RELATED WORKS}
\label{section:related-works}

Following the prevailing camera calibration methods, many works have implemented lidar-camera calibration using a planar checkerboard. Zang et al.~\cite{Zhang2004} use a checkerboard with a two step estimation approach to calibrate one Red-Green-Blue (RGB) camera to one 2D lidar. Kim et al.~\cite{Kim2019} use a checkerboard to calibrate a 16 beam 3D lidar with six monocular cameras, where each camera-lidar pair is treated as its own estimation problem. In both cases, the targets must be moved around to collect measurements of the target at different view points. In the work by Geiger et al.~\cite{Geiger2012}, multiple checkerboards are used to calibrate any number of cameras and lidars using a single measurement of the checkerboards. Similarly, Alismail et al.~\cite{Alismail2012} use a planar target with a single black circle that can be detected in the image which is used to calibrate a rotating 2D lidar to a camera.

In the above cases, distinct features on the target cannot be detected in the lidar data and therefore the overall shape of the target is used with iterative correspondence methods or estimated plane properties. To avoid this issue and improve the accuracy of calibrations, other groups have designed targets with distinct detectable features. Rodriguez et al.~\cite{RodriguezF.2008} use a planar target with a single hollowed circle which allows the center and radius of the circle to be measured and used for optimization. Both Velas et al.~\cite{VelasM2014} and Guindel et al.~\cite{Guindel2018} use a planar target with 4 hollowed circles and the centers of the circles are extracted from both the image data and the lidar data to be used as features. 

In all the above cases, care has to be taken to place the targets within the FOV of the sensors to ensure sufficient scan points have contacted the targets, and the target is fully visible in the camera. This limits the possible sensor configurations to those with sufficient overlap, and the calibration results can degrade as the overlap becomes smaller due to a reduced measurement sample space~\cite{Rebello2020}.

Thus far, all the aforementioned works are target-based calibration approaches which have been implemented to estimate one lidar to one camera extrinsic calibration. For systems with multiple lidars and/or cameras, these calibrations would need to be repeated for every camera-lidar pair. This can lead to a build-up of error if a chain of transformations is required and each link is estimated independently from one another. Domhof et al.~\cite{Domhof2019} solve this problem by performing global optimizations of multiple cameras and lidars. This methodology also uses a planar target with four circular holes, while extending the approach to radar extrinsic calibration using a protruding triangle in the center. Kummerle et al.~\cite{Kummerle2018} use a large sphere as a target where the center of the sphere is detected in the image and lidar measurements. This method also performs a global optimization to estimate the extrinsic transformations for any number of cameras and lidar which eliminates the possibility of error accumulation with a pairwise extrinsic calibration method. These methods, however, still require overlap in the sensor FOVs, and the benefits of the global optimization of all calibrations may not be beneficial if the overlap is restricted to a few sensor pairs.

Targetless methods seek to define a calibration approach that does not explicitly rely on a particular calibration target, enabling calibration in the field and online calibration during robot operation. To do so usually requires robust correspondence of features or maps between lidar and camera data of some form. In the case of Castorena et al.~\cite{Castorena2016} and Levinson et al.~\cite{Levinson2013}, targetless calibration is performed by associating depth discontinuities in lidar data with intensity discontinuities in image data. This approach can be problematic because color discontinuities do not always match with range discontinuities. Pandey et al.~\cite{Pandey2012} and Taylor et al.~\cite{Taylor2013} maximize mutual information in both the lidar data and camera data, but this method suffers from incorrect correspondences when initial calibration estimates are insufficient~\cite{Kummerle2018}. Repeatable accurate calibrations can be challenging in these methods contrary to methods performed in a lab setting with a known calibration environment having distinct features. Yuan et al.~\cite{yuan2021pixel} use edge-based targetless approaches for extrinsic calibration between camera and lidar, but heavily rely on scenes rich in geometric features. Koide et al.~\cite{koide2023general} proposed a general targetless lidar-camera calibration toolbox, but still required significant sensor FOV overlap for reliable results. However, these approaches can be well suited for checking for de-calibration while in operation after a more reliable initial calibration has been performed. So far, the above targetless calibration methods still do not solve the problem of requiring significant overlap in the sensor FOVs. 

Two general approaches have been proposed for non-overlapping targetless calibration: (1) SLAM is performed while the sensors are in motion and the calibrations are optimized (~\cite{Scott2015}, ~\cite{Napier2013},~\cite{9779777}), or (2) motion estimates are constructed with each sensor data stream individually, and a hand-eye calibration is performed to align the motion estimates offline (~\cite{Taylor2016},~\cite{Andreff2001},~\cite{Heller2011}). Since SLAM can be very environment dependent, these calibration approaches are limited by the appearance and geometric properties of the environment, which means calibration accuracy cannot be guaranteed. Jeong et al.~\cite{Jeong2019} also solve this problem for a lidar and stereo camera pair by using lane markings in roads, but this is limited to those applications for which such markings are available.

\section{METHODOLOGY}
\label{section:meth}

\subsection{Problem Formulation and Notation}
\label{section:meth-problem}

In this section, we will present the mathematical formulation of the problem and notation for a calibration setup with one lidar, one camera, and one target. It is important to highlight that this formulation can be applied to any number of cameras, lidar, and targets in your setup, and the released open source tooling has such an implementation. Nonetheless, we simplify our description herein for brevity. 

\begin{figure}[thpb]
  \centering
  \includegraphics[width=0.25\textwidth]{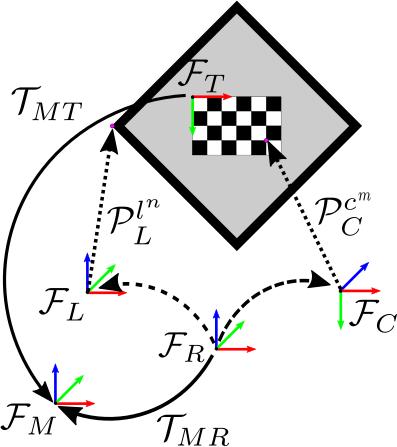}
  \caption{Calibration problem with all coordinate frames}
  \label{fig:coordinate_frames}
\end{figure}

Fig. \ref{fig:coordinate_frames} shows a typical calibration scenario and illustrates all frames and transformations. Let $F_{L}$, $F_{C}$, and $F_{T}$ be the coordinate frames for the lidar, camera, and target, respectively. $F_M$ is the stationary map frame in which all MCS measurements are expressed. Finally, a robot frame, $F_R$, is defined as the sensor base frame tracked by the MCS and can be arbitrarily located on the robot. 

There are three types of quantities that are measured and estimated. The first quantity, denoted by dotted lines on Fig. \ref{fig:coordinate_frames}, are the keypoints on the target as measured from the sensor frames. For the case of the lidar, $P^{l^n}_{L}$ is the measured vector from $F_{L}$ to the lidar keypoint $l^n$. Similarly, $P^{c^n}_{C}$ is the measured ray (with unknown depth) from $F_{C}$ to the camera keypoint $c^n$. The second quantity, denoted by solid lines, are transformations measured by the MCS. Let $T_{MT}$ be the rigid transformation that can be used to transform points from the target frame to the map frame. Similarly, let $T_{MR}$ be the transformation from the robot frame to the map frame. Both these transformations are estimated at a high rate relative to the sensor data rate, and to a sub-mm level of accuracy~\cite{Merriaux2017}. The third quantity, denoted by dashed lines on Fig. \ref{fig:coordinate_frames}, are the calibrations that are being estimated. In this case, the unknown calibrations are from $F_R$ to $F_{C}$ and from $F_R$ to $F_{L}$. 

If initial estimates of the calibrations (dashed lines) are available, all frame poses are known at all times, forming a closed loop for each measurement. In our formulation, a measurement is a keypoint observation from a camera or lidar at a specific time instance. Since the MCS measurement frequency can be an order of magnitude greater than that of the lidar and camera data, all sensor measurements should have an associated MCS measurement very close in time, thus interpolation error can be minimized without explicit hardware synchronization of sensor and MCS capture. This ensures that measurements can be taken reliably while the robot and/or the targets are in motion, and without the need to integrate timing circuitry between the MCS and robotic system. Since the data for most lidar is continuous, if the lidar is in motion, it is expected that the lidar points are motion compensated to discrete time intervals. The MCS and initial calibration estimates can be used for lidar motion compensation.

\subsection{Mathematical Framework}
\label{section:meth-math}

Let us first consider the case of extrinsic lidar calibration to the robot frame as a separate problem. The problem can be written as follows: given a set of measured lidar keypoints, $\mathbf{P}^{l}_{L} = \{ P^{l^1}_{L}, \dots , P^{l^N}_{L}\}$, an initial calibration estimate, $T_{RL}$, as well as the set of corresponding ground truth keypoint locations expressed in the robot frame, $\mathbf{P}^{l}_{R} = \{ P^{l^1}_{R}, \dots , P^{l^N}_{R}\}$, estimate the robot to lidar frame transformation parameters. In this case, $N$ is the total number of lidar keypoints which depends on the number of target observations and the number of keypoints per target observation. To determine the ground truth location of the keypoints in the robot frame, we can use the location of the keypoints in the target frame (i.e., $P_T^{l^i}$) which are known a priori (e.g, hand measured or known due to target dimensions) and the high precision MCS measurements according to Eq. \ref{eqn:lidar_keypoint_transforms}. For details on how measured lidar keypoints are extracted and how their corresponding ground truth values are determined, see Section \ref{section:meth-corr-opt}.

\begin{equation}
\label{eqn:lidar_keypoint_transforms}
    P^{l^n}_R = T_{RM} \: T_{MT} \:  P^{l^n}_T
\end{equation}

Using Eq. \ref{eqn:lidar_keypoint_transforms} and the initial calibration estimate, an error associated with the measurement of each lidar keypoint can be defined as a function of the calibration, as shown in Eq. \ref{eqn:lidar_keypoint_error}. This error represents the Euclidean distance between the detected lidar point, expressed in the lidar frame, and its corresponding ground truth lidar point transformed to the lidar frame using the current calibration estimate. As the current calibration estimate converges to the true value, this error should diminish. 

\begin{equation}
\label{eqn:lidar_keypoint_error}
    e_{l,n}(T_{RL}) = P^{l^n}_L - T_{RL}^{-1} \: T_{RM} \: T_{MT} \: P^{l^n}_T
\end{equation}


Let us now consider the case of camera extrinsic calibration. Given a set of measured camera keypoints, $\mathbf{\rho}^{c} = \{ \rho^{c^1}, \dots , \rho^{c^M}\}$, an initial calibration estimate, $T_{RC}$, as well as the set of ground truth keypoint locations expressed in the robot frame, $\mathbf{P}^{c}_{R} = \{ P^{c^1}_{R}, \dots , P^{c^M}_{R}\}$, estimate the true calibration. Similar to the lidar case, an error term can be defined using the MCS measurements and the initial calibration. In this case, the camera-specific projection function $\pi(\dotsc)$ can be used to project the estimate of the ground truth keypoint location into the image plane. Eq. \ref{eqn:camera_keypoint_error} shows this error function which represents the difference between the detected pixels and the projection of their corresponding keypoints after they have passed through the MCS transforms and the current calibration estimate. Note that in this work, we assume the intrinsics have already been calibrated, however, expanding this implementation to optimize over intrinsics is a straightforward extension that will be left for future work. 

\begin{equation}
\label{eqn:camera_keypoint_error}
    e_{c,n}(T_{RC}) = \rho^{c^m} - \pi(T_{RC}^{-1} \: T_{RM} \: T_{MT} \: P^{c^m}_T)
\end{equation}

In practice, we cannot exactly determine the keypoint locations relative to the target frame that is measured by the MCS. That is because the transform from the keypoint frame to the target frame often has to be hand measured or manufacturing defects may cause the transform to be slightly different than designed. Therefore, to account for this error, we add an additional target alignment error term, $T_E$, for which we also optimize. $T_E$ essentially represents a transformation from the estimated target frame to the corrected target frame. This has been shown to significantly increase the accuracy of the final results in Sec.~\ref{sec:ablation_study}. Including these target error parameters yields Eqns. \ref{eqn:lidar_keypoint_error_final} \& \ref{eqn:camera_keypoint_error_final}.

\begin{equation}
\label{eqn:lidar_keypoint_error_final}
    e_{l,n}(T_{RL}) = P^{l^n}_L - T_{RL}^{-1} \: T_{RM} \: T_{MT} \: T_{E} \: P^{l^n}_T
\end{equation}

\begin{equation}
\label{eqn:camera_keypoint_error_final}
    e_{c,m}(T_{RC}) = \rho^{c^m} - \pi(T_{RC}^{-1} \: T_{RM} \: T_{MT} \: T_{E} \: P^{c^m}_T)
\end{equation}

To solve for the calibrations, we use a non-linear solver to find the set of $T_{RL}$, $T_{RC}$, and $T_{E}$ calibrations that minimize the sum of the error terms for camera and lidar cost functions. Since all errors are added to a single optimization problem, and initial target location estimates are known, targets can be used that cannot fully constrain the object pose from a single detection of that target, as is shown in Section \ref{section:results-sim} with a cylindrical target.


\subsection{Measurement Extraction}
\label{section:meth-measurements}

\begin{figure}[thpb]
  \centering
  \includegraphics[width=0.48\textwidth]{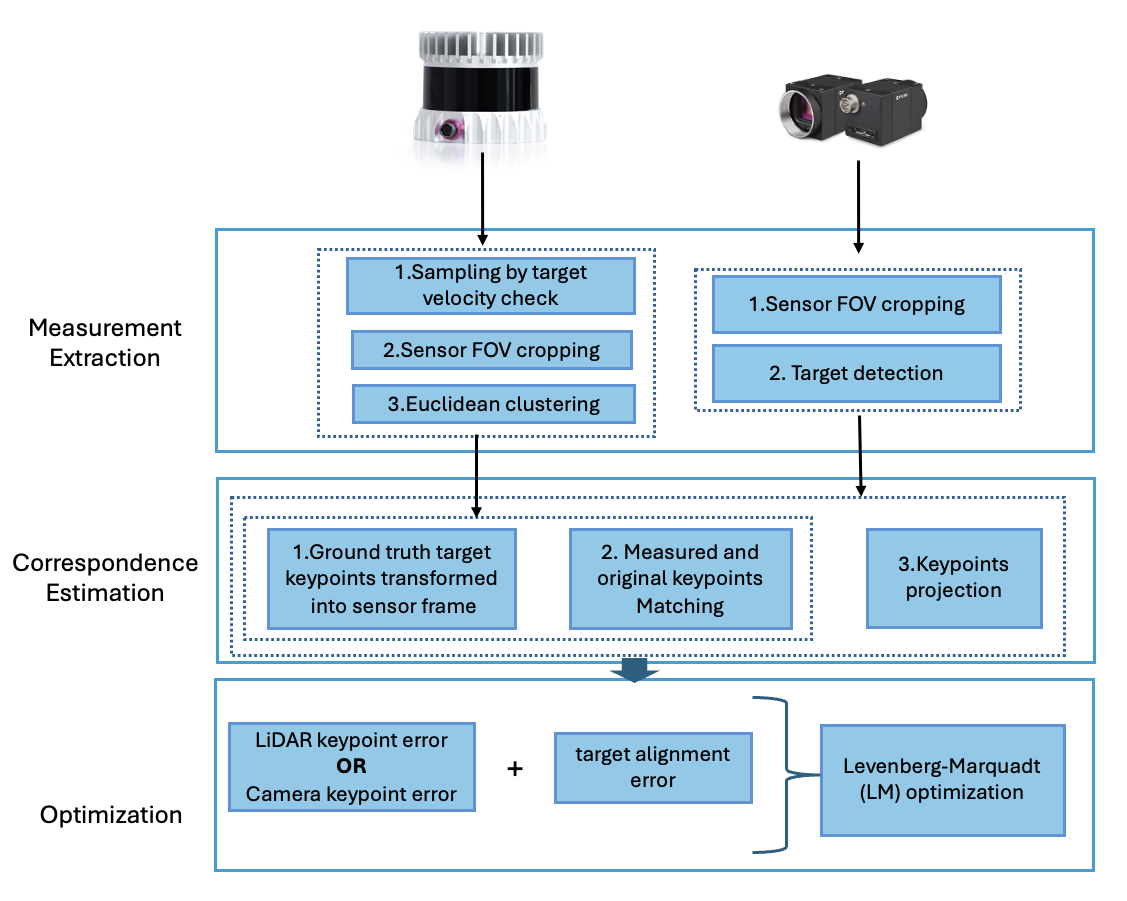}
  \caption{Calibration process overview}
  \label{fig:method_flowchart}
\end{figure}

As discussed in the introduction, a key novelty of this work is that we formulate the problem in a way that is agnostic to the target being used, and to the method for extracting measured keypoints. We use the fact that with initial estimates of the sensor extrinsics, we have a fully closed loop of transformations so we can transform measured keypoints and their corresponding known keypoints into a consistent reference frame and find the corresponding features using nearest neighbors. For more information on correspondence estimation, refer to Section \ref{section:meth-corr-opt}. Using a point-to-point error term in the optimizer then makes this formulation agnostic to any target design. Furthermore, we add all point-to-point measurements to a single optimization problem, meaning that each observation of a target does not need to constrain all 6DOF. Instead we rely on multiple measurements of possibly degenerate target measurement to constrain all calibration DOFs. As a result, this work presents the first implementation that can be used by researchers to experiment on a large variety of different target designs and feature extraction methods. The calibration process is illustrated in Fig.~\ref{fig:method_flowchart}.

For our target design, we select a diamond shaped board with a center checkerboard, while also demonstrating another target design option in simulation which is presented in Section \ref{section:results-sim}. The diamond shape provides a large surface area for the lidar to scan, with long distinct edges. The center checkerboard allows robust keypoints to be extracted from camera images. For extracting lidar measurements, we experiment with a method for estimating the four corner points of the diamond, as well as using all points that come into contact with the target. We determined that the latter provided us with better calibration results. For the checkerboard detection, we simply use OpenCVs built-in functions. 


Some data pre-processing steps can be taken to decrease the probability of using outlier target measurements which are only possible due to the unique circumstances offered by the MCS (i.e., the pose of the target is always approximately known). First, we step through our dataset and at each time step we calculate if the target is expected to be within the FOV of each sensor. Next, we check the velocity of the target relative to the robot and reject measurements with high motion which may induce motion distortion or may amplify the effect of any time synchronization errors. Once we know a target is likely to be observed by a sensor and is sufficiently stationary, we crop the FOV of the sensor to capture only an area surrounding the estimated target location. After cropping the lidar data to the expected location of the target, we further narrow down the points that we expect to be part of the target by performing Euclidean clustering and selecting the most likely cluster. The cluster tolerance can be predetermined for each measurement given the known minimal angular resolution of the lidar and the known distance to the target. The most likely cluster is selected based on a weighted score using the difference between the target pose and the measured cluster error, and the difference between the measured volume and the known target volume. This results in a small subset of the collected data for both sensor types in which to identify the target, which increases our likelihood of properly detecting the correct keypoints on the target. We argue that these data pre-processing steps enable our method to have much more robust measurements compared to existing methods.

\subsection{Correspondence Estimation and Optimization}
\label{section:meth-corr-opt}

To construct the error terms presented in Eqns. \ref{eqn:camera_keypoint_error_final} \& \ref{eqn:lidar_keypoint_error_final}, we must associate keypoint measurements in the sensor frame ($P_L^{l^n}, \rho^{c^n}$), to the ground truth keypoint locations. Since each keypoint is not always uniquely identifiable, we estimate the correspondences between the measured points and the ground truth keypoints iteratively. 

\begin{figure}[h]
\begin{center}
\begin{tabular}{c c}
     \includegraphics[width=0.2\textwidth]{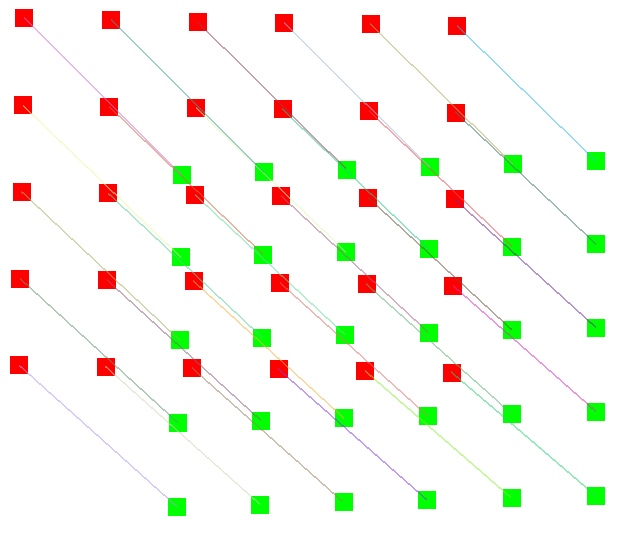} &  \includegraphics[width=0.2\textwidth]{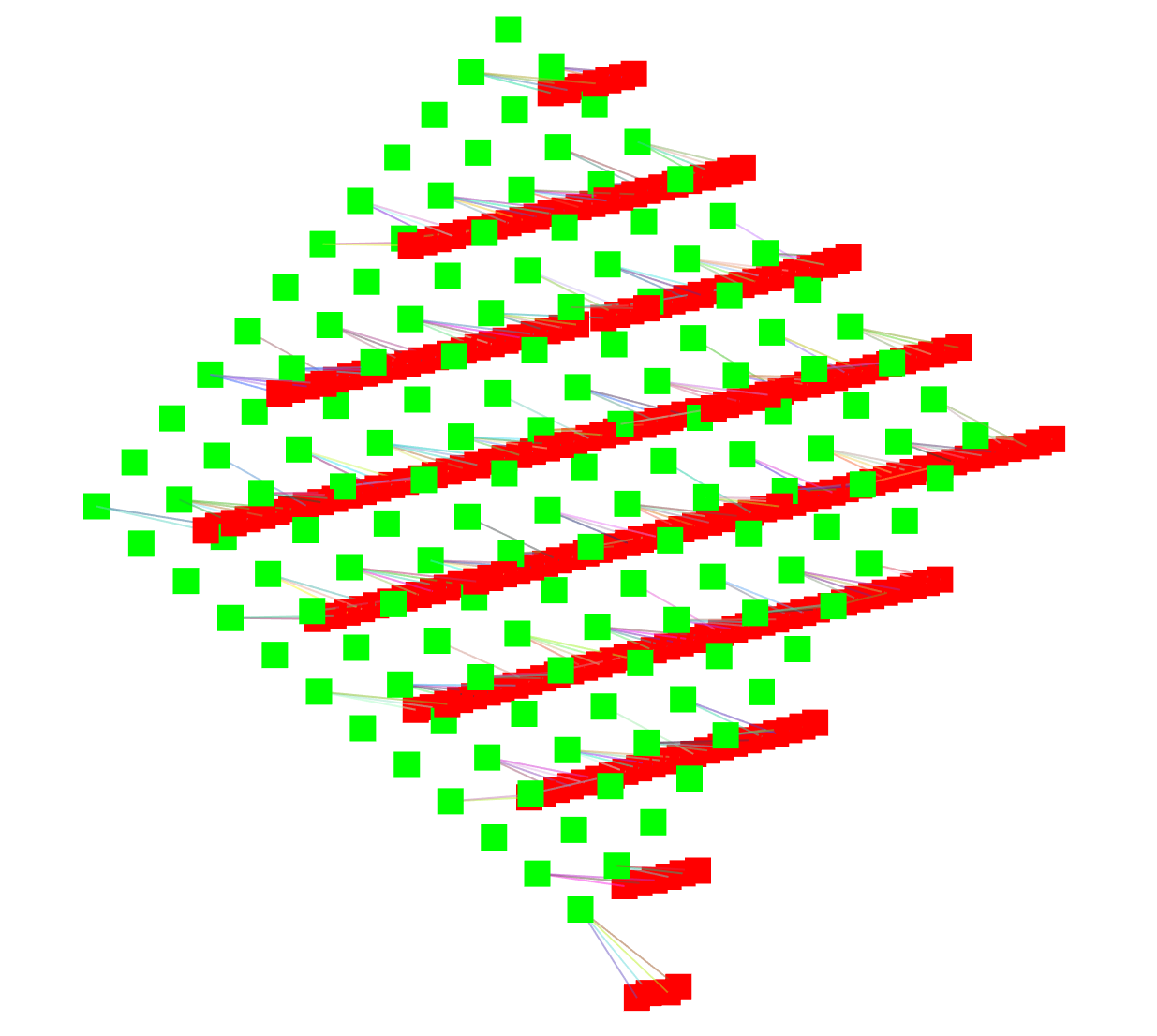}\\
\end{tabular}
\end{center}
\caption{Example correspondence estimation for diamond target: camera keypoints (left), lidar keypoints (right). Red: measured keypoints. Green: transformed ground truth keypoints}
\label{fig:correspondences}
\end{figure}

To solve this correspondence problem for a given target, we can take advantage of the MCS measurements and the current estimate of the calibration to perform a nearest neighbor search (NNS). First, the ground truth target keypoints are transformed from their target frames into the sensor frames. Following this step, the centroids of both point clouds are calculated, and the transformed keypoints are translated such that the centroids of both clouds align. A NNS is then performed to determine the correspondences between the measured keypoints and the original keypoints on the target. For the case of the camera measurements, the keypoints are also projected into the image plane and NNS is done on the 2D pixel coordinates. Fig. \ref{fig:correspondences} shows an example of the camera and lidar correspondence estimation for the diamond target.

The optimization of the presented cost function is solved using Levenberg-Marquadt (LM) optimization and implemented with Ceres\footnote{Ceres is available at: http://ceres-solver.org/}. Since the original correspondences are often incorrect due to error in the initial calibrations used to transform the keypoints, the correspondence search and optimization are iterated until convergence. That is, once the optimizer converges, the correspondences are recalculated and a new optimization problem is constructed and solved. This process is iterated until either the change in total cost or the change in calibration parameters is below a threshold.

\section{RESULTS}
\label{section:results}

This section will discuss the validation procedure for the proposed calibration approach and present key results. The results will be presented for simulation and experimental data where simulation is used to test against known calibrations that have been perturbed. Both simulation and experimental data are validated using average keypoint reprojection error for the case of the camera calibrations, and average keypoint Euclidean distance error for the lidar calibrations. 

\subsection{Simulation Data}
\label{section:results-sim}

A simulation environment was created using Gazebo and the Robot Operating System (ROS) to simulate a MCS calibration environment with one camera and one Velodyne VLP-16 lidar. To simulate the camera, a simple pinhole camera with no lens distortion is used to avoid any possible issues with improper intrinsic calibrations. Gaussian noise was added to the pixel intensities measured by the simulated camera with a mean of 0 and standard deviation of 1.8. The lidar was simulated using Velodyne's ROS simulator. To simulate the MCS, a ROS node was created that converts the model states provided by Gazebo to transform messages that describe the target motion relative to the map frame, and the robot frame motion relative to the map frame. These odometry messages have been designed to emulate the data format from a MCS. No measurement noise is added to the MCS measurements due to the high accuracy that these devices can achieve relative to the accuracy of the sensors being calibrated. The Vicon MCS used in this work can achieve a mean absolute error of 0.15 mm~\cite{Merriaux2017}, which is orders of magnitude smaller than, for example, typical lidar error which are on the order of several centimeters. 

For simulation data, the true calibrations are known, therefore the rotational and translational errors of the calibrations (i.e., transformations from sensor to robot frame) are presented. For the rotational error, the angles are converted to angle-axis representations and the absolute difference in the angle is presented. For the translational error, the absolute difference in the translation norms is presented. For the case of the camera calibration, the reprojection error is also presented by taking the norm of the vector between the projected points and the detected pixels. For the case of the lidar, the mean Euclidean error is presented which is calculated by taking the norm of the vector between the target keypoints transformed into the lidar frame and the detected keypoints in the lidar scan data. To validate the sensitivity of the proposed approach towards various initial calibration estimates, the known calibrations are perturbed by sampling linear and rotational perturbations from a random uniform distribution. For translational DOFs, perturbations are taken from -3 cm to +3 cm, and for rotational DOFs, perturbations are taken from -5 degrees to +5 degrees. The maximum perturbations are justified based on the maximum error that could be expected by hand measuring transformation between sensor centers.

Tables \ref{table:results-sim-lid-dia} show the lidar calibration results using the diamond target. Tests were done with 5, 15, and 30 measurements of the target at different positions and orientations. In general, the mean and standard deviations of the sensor extrinsics and Euclidean errors are reduced as the number of measurements increases, but convergence occurs even with few measurements. Looking at the diamond target results, a mean of 0.3 mm of translational error and 0.004 degrees of rotational error in the extrinsics is achieved after only 5 measurements. 

\begin{table}[h]
\caption{Simulated lidar calibration errors for the 3 tests}
\label{table:results-sim-lid-dia}
\centering
\begin{tabular}{cccccccc}
\hline
No. & \multicolumn{2}{c}{Trans. (mm)} & \multicolumn{2}{c}{Rot. (deg)} & \multicolumn{2}{c}{Euc. (mm)}\\
Mea. & $\mu$ & $\sigma$ & $\mu$ & $\sigma$ & $\mu$ & $\sigma$ \\
\hline
30 & 0.1 & 1.7e-6 & 1.8e-3 & 5.1e-8 & 0.5 & 0.1 \\ 
15 & 0.1 & 2.0e-6 & 2.4e-3 & 3.6e-7 & 0.4 & 0.6 \\ 
5  & 0.3 & 4.3e-6 & 3.8e-3 & 3.5e-8 & 2.2 & 9.0 \\   
\hline
\end{tabular}
\end{table}


Table \ref{table:results-sim-cam-dia} presents the results of camera calibration using the diamond target. Similar to the lidar calibration results, the camera calibration converges with only a few measurements. After 5 measurements, the mean translation error and rotational error are 0.1 mm and 0.03 degrees, respectively. The reduction in the errors as the number of measurements increases is less apparent in this case. This is likely because the calibration errors are so low with only 5 measurements, and increasing the number of measurements introduces the possibility of increased error due to motion blur or other such error sources.

\begin{table}[h]
\vspace{0.1cm}
\caption{Simulated camera calibration errors for the 3 tests}
\label{table:results-sim-cam-dia}
\centering
\begin{tabular}{cccccccc}
\hline
No. & \multicolumn{2}{c}{Trans. (mm)} & \multicolumn{2}{c}{Rot. (deg)} & \multicolumn{2}{c}{Proj. (pixels)}\\
Mea. & $\mu$ & $\sigma$ & $\mu$ & $\sigma$ & $\mu$ & $\sigma$ \\
\hline
30 & 0.111 & 1.3e-6 & 0.035 & 9.2e-8 & 0.158 & 1.4e-8 \\ 
15 & 0.066 & 3.6e-6 & 0.035 & 3.6e-6 & 0.167 & 6.2e-8 \\ 
5  & 0.136 & 1.9e-6 & 0.034 & 1.3e-8 & 0.131 & 3.9e-7 \\   
\hline
\end{tabular}
\end{table}

\begin{figure}[h]
\centering
  \begin{subfigure}[b]{0.2\textwidth}
    \includegraphics[trim={0cm 1cm 0cm 1cm},clip, width=\textwidth]{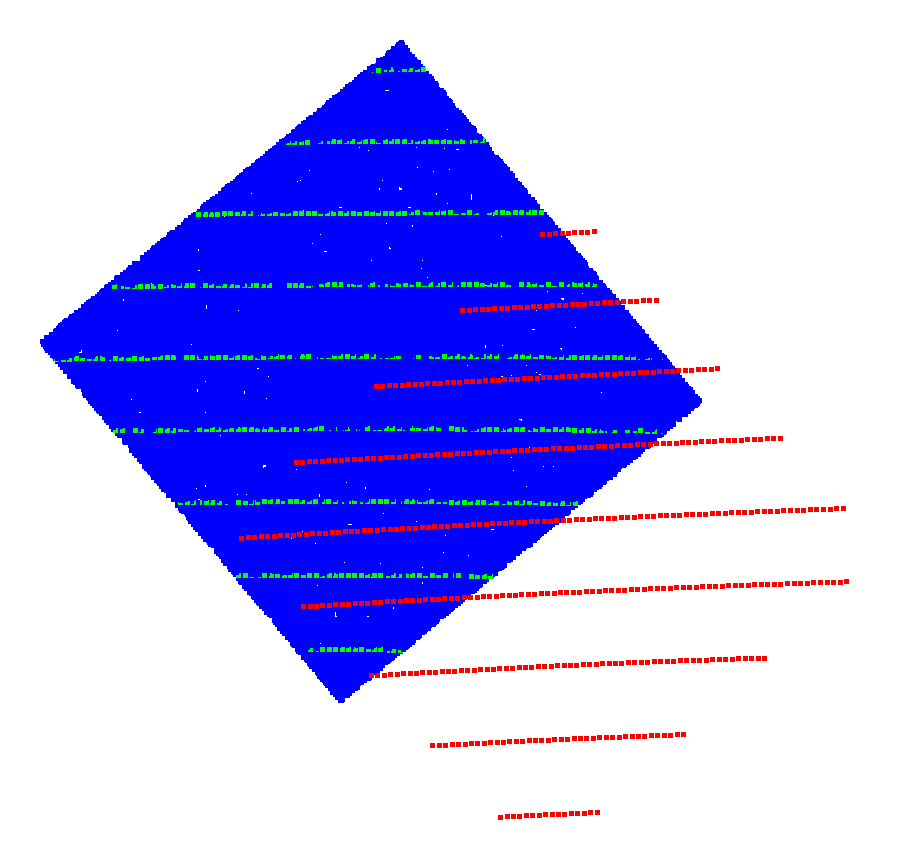}
  \end{subfigure}
  \begin{subfigure}[b]{0.2\textwidth}
    \includegraphics[trim={18cm 18cm 18cm 0cm},clip, width=\textwidth]{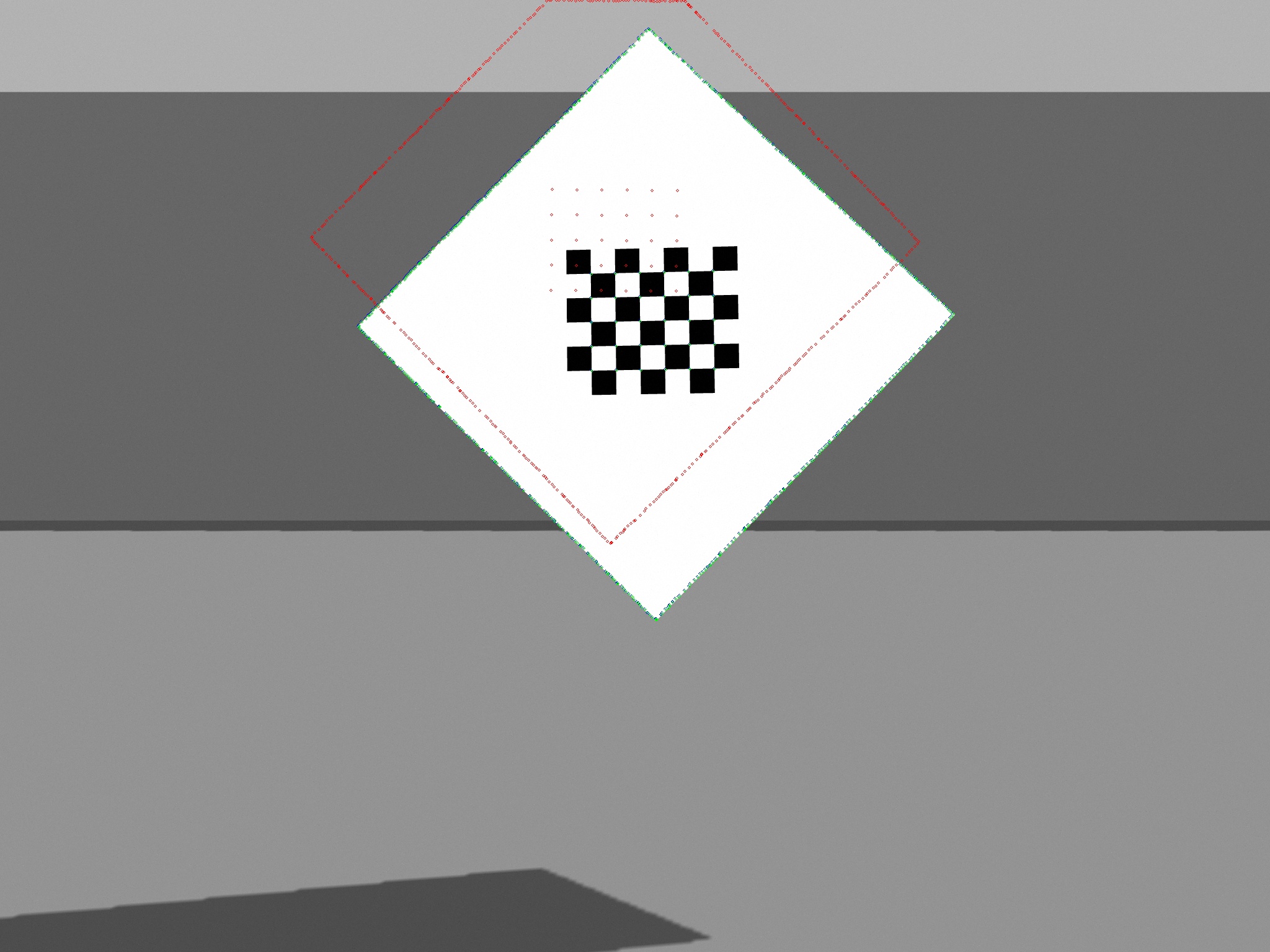}
  \end{subfigure}
\caption{Example lidar calibration (left) and camera calibration (right) results using simulation data with diamond target}
\label{fig:results-lid}
\end{figure}

To illustrate the calibration results, we can conveniently use the pose of the target at each measurement from the MCS to overlay results. Given the template cloud for each target, we can transform the template cloud points and the measured keypoints into the map frame using the initial, optimized, and ground truth calibration to see the accuracy of our calibration. Figure \ref{fig:results-lid} shows one measurement example for each lidar target, where the blue points are the template cloud points transformed into the map frame using the ground truth calibration (i.e., the ground truth target pose), the green points are the scan points transformed to the map frame using the optimized calibrations, and the red points are the scan points converted to the map frame using the initial calibration (i.e., the perturbed ground truth calibration). The green points line up very well with the blue template cloud which visually demonstrates the accuracy of our calibration.

Figure \ref{fig:results-lid} shows similar results for the camera calibration, however, the points are projected into the image plane and plotted on the image measurement for visualization. In this case, the keypoints and the outline of the diamond target are shown to better visualize the results. 

\begin{figure}[h]
\centering
  \begin{subfigure}[b]{0.2\textwidth}
    \includegraphics[trim={0cm 1cm 0cm 1cm},clip, width=\textwidth]{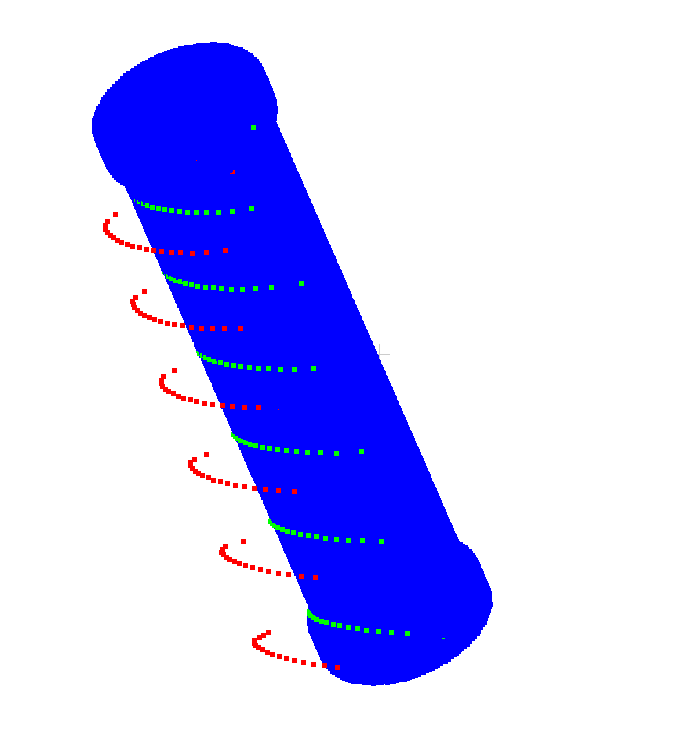}
  \end{subfigure}
  \begin{subfigure}[b]{0.2\textwidth}
    \includegraphics[trim={15cm 5cm 15cm 7cm},clip, width=\textwidth]{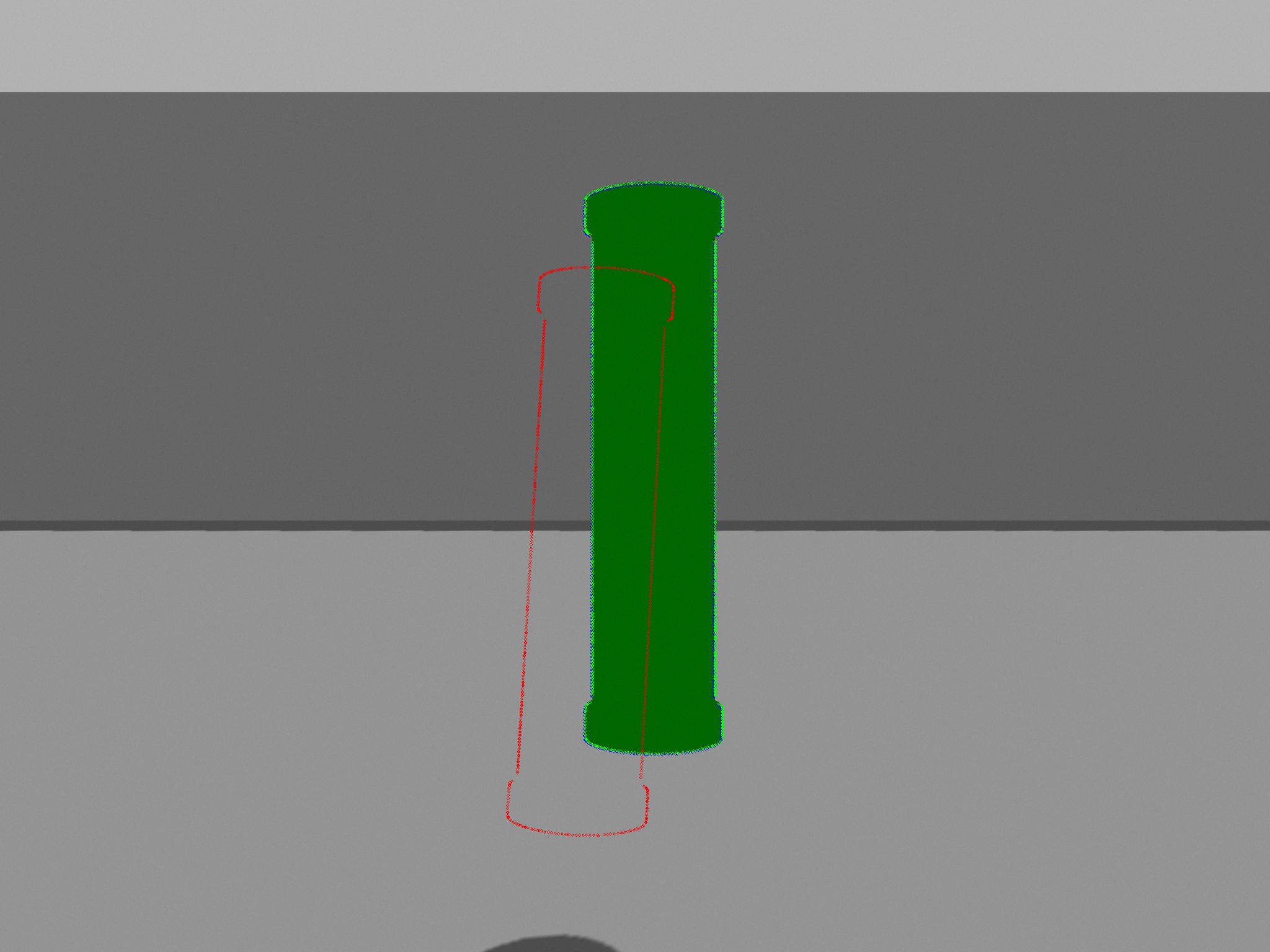}
  \end{subfigure}
\caption{Sample results using cylinder target for: lidar calibration (left) and camera calibration (right)}
\label{fig:results-cylinder}
\end{figure}

To demonstrate that the approach can be used for various target designs, including targets that are normally considered as degenerate on their own, we also present results using a cylinder target. Fig. \ref{fig:results-cylinder} shows example calibration results using this cylinder. For extracting lidar keypoints, we use all lidar points that contact the cylinder. For camera keypoint extraction, we first use a color thresholding to extract the target from the image, and then extract the convex hull from those points. Since this is a cylinder, a single measurement of the target would present a degenerate case for lidar and camera since the rotational axis of the target is ambiguous. However, using multiple measurements of the target, we can still accurately recover the calibrations. In these tests, with only 15 measurements, we achieve a calibration errors of 0.1 mm \& 0.002 deg for lidar calibration, and 4.6 mm \& 0.1 deg error for camera calibration.

\subsection{Experimental Data}
\label{section:results-experimental}

\begin{figure}[h]
\centering
  \begin{subfigure}[b]{0.19\textwidth}
    \includegraphics[trim={0cm 0cm 0cm 0cm},clip, width=\textwidth]{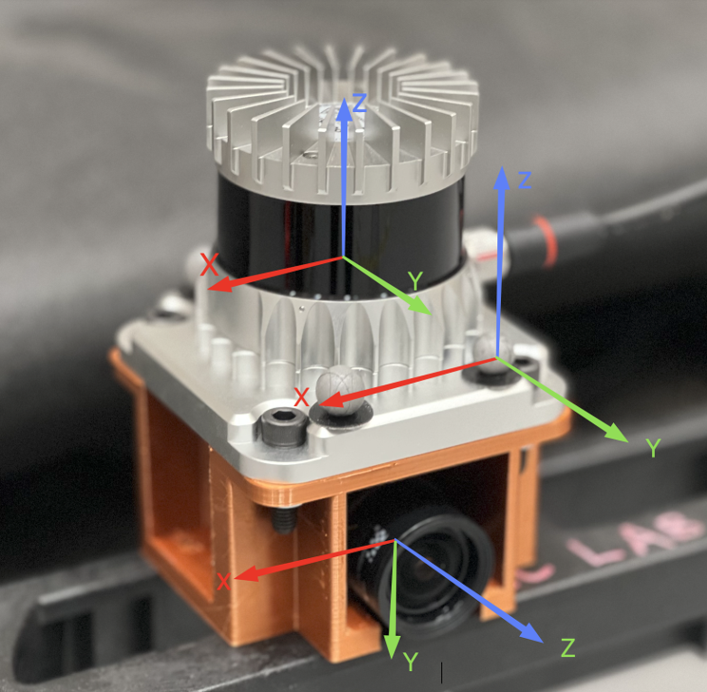}
  \end{subfigure}
  \begin{subfigure}[b]{0.2\textwidth}
    \includegraphics[trim={0cm 0cm 0cm 0cm},clip, width=\textwidth]{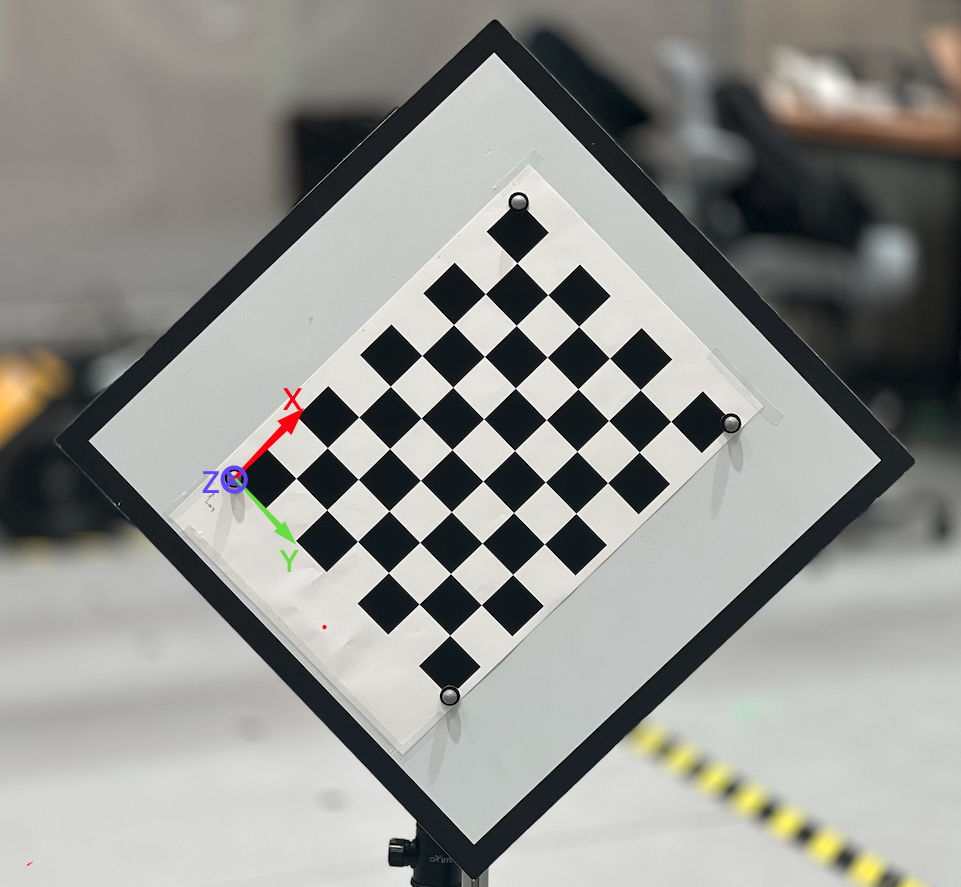}
  \end{subfigure}
\caption{Experimental sensor setup (left) and target setup (right)}
\label{fig:ouster-lidar}
\end{figure}

To validate our methodology in the real world, we collected several datasets using a single lidar and camera rigidly mounted to a hand-held device, as shown in Fig. \ref{fig:ouster-lidar}. In this configuration, the lidar and camera do have highly overlapping FOVs, however, as described in Section \ref{section:meth}, all lidar and camera frames get calibrated relative to the MCS frame on the sensor rig. Therefore, we do not need to prove experimentally that non-overlapping FOV sensors can be calibrated since we know mathematically that we can always retrieve the relative transforms between all cameras regardless of their viewing angles relative to each other.

The lidar is an Ouster OS0-32 Rev6 which is a 32 beam mechanically rotating lidar. The camera is a Flir Blackfly camera (BFS-PGE-16S2C-CS), with $720\times540$ resolution and a $49\times38$ degree FOV lens. The camera is intrinsically calibrated using the Matlab camera calibration toolbox with a pinhole camera model with radial-tangential lens distortion. The MCS used in the experimental data is a Vicon MCS~\cite{Merriaux2017}, and markers were setup on the targets (as shown in Fig \ref{fig:ouster-lidar}) so that the MCS can directly measure the pose of the targets. All keypoint locations were hand measured a priori in these coordinate frames, and template clouds were generated with the coordinate centers aligning with that of the real targets. MCS markers were added to the sensor rig, ensuring good visibility by the MCS cameras. For convenience of use of checkerboard template, coordinate frame of the target is set parallel to the diamond board in x and y axes. MCS markers can be precisely placed along the axis of the diamond target. The location of the robot coordinate frame is irrelevant, since all calibrations are measured with respect to the same frame and therefore relative calibrations between all sensors can be determined.

We use the MSC measurements to extract measurements when target and sensors fixed at discrete locations. In both camera and lidar datasets, at least 20 measurements of the target at different positions and orientations are collected. Extrinsics are initialized using rough hand measurements and visually verified in rviz. 

\begin{figure}[h]
\centering
\includegraphics[trim={0.5cm 2cm 1cm 2cm},clip, width=0.4\textwidth]{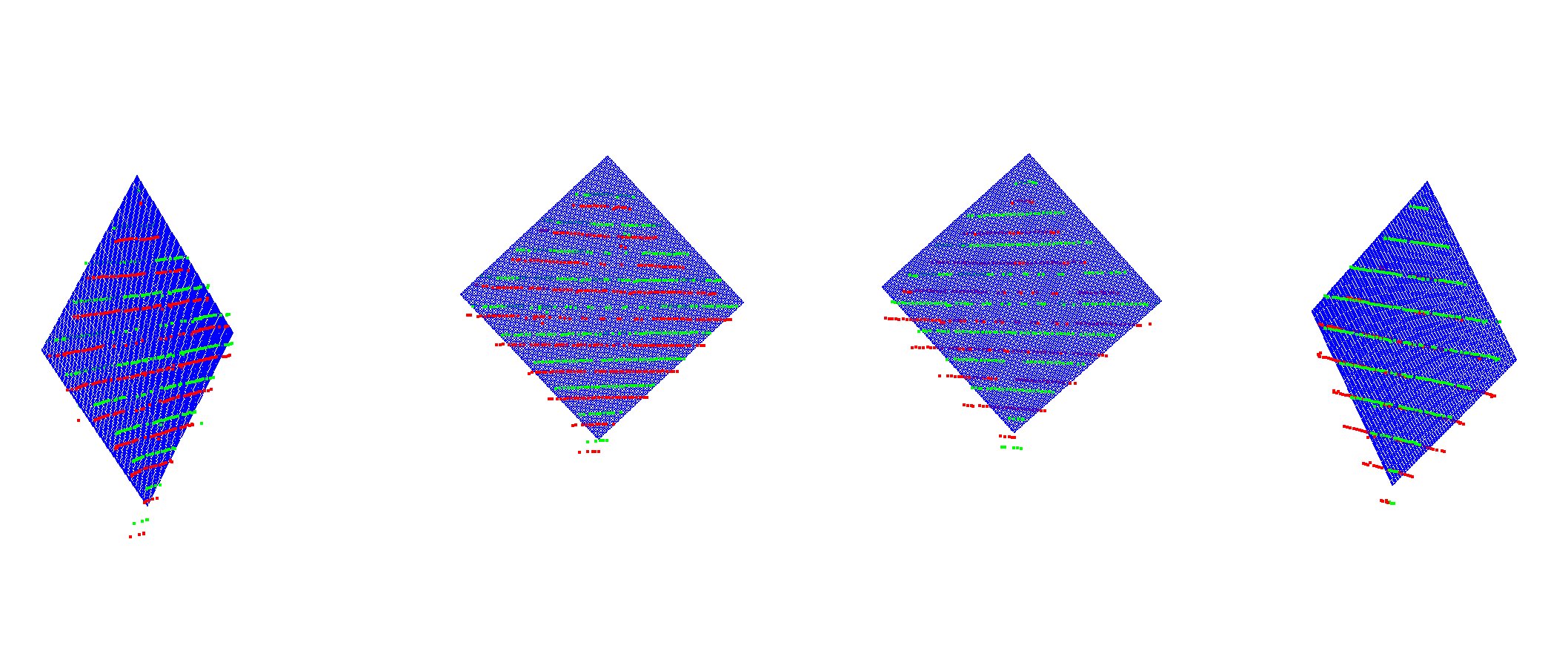} \\
\caption{ 4 sample lidar results}
\label{fig:results-real-lid}
\end{figure}

Since the simulation results suggest that the calibrations should converge with at least 15 target measurements, 15 to 23 measurements were used for experimental testing in Table \ref{table:results-real-dia}, where calibration results converge and are consistent across measurements. For lidar calibration using the diamond target, an average Euclidean error of 0.016m was achieved. According to the manufacturer's datasheet, the OS0-32 lidar shows a typical range error of +/-1.5 cm, suggesting that the +/- error of 1.6cm to 1.7cm is well within the error limits that would be expected even with a perfect calibration. This demonstrates promising results for the lidar calibration given the high level of measurement noise in the sensor data.

Fig \ref{fig:results-real-lid} shows an example of lidar calibration results on the experimental data. Once again, the blue points are the template clouds converted into the map frame given the MCS measurements. The green and red points are the scans points converted into the map frame using the optimized and initial calibrations, respectively. Despite the high noise in the lidar data, the green points visually align well with the blue template cloud. More lidar measurements in different angles and distances are also shown in Fig \ref{fig:results-real-lid}.

\begin{table}[h]
\vspace{0.1cm}
\caption{Experiment calibration errors}
\label{table:results-real-dia}
\centering
\begin{tabular}{c c c }
\hline
Lidar Test & No. Meas. & Trans. (m)  \\
\hline
1 & 23 & 0.0162282 \\
2 & 20 & 0.0164762 \\ 
3 & 15 & 0.0167934 \\   
\hline
\end{tabular}

\vspace{1em}

\begin{tabular}{cccccccc}
\hline
Camera Test & No. Meas. & Proj. (deg) & Proj. (pixels) \\
\hline
1 & 20   & 0.04273 & 0.72292  \\ 
2 & 15   & 0.04097 & 0.69323  \\ 
3 & 10   & 0.03142 & 0.53167  \\   
\hline
\end{tabular}

\end{table}

\begin{figure}[h]
\centering
  \begin{subfigure}[b]{0.16\textwidth}
    \includegraphics[trim={0cm 0cm 0cm 0cm},clip, width=\textwidth]{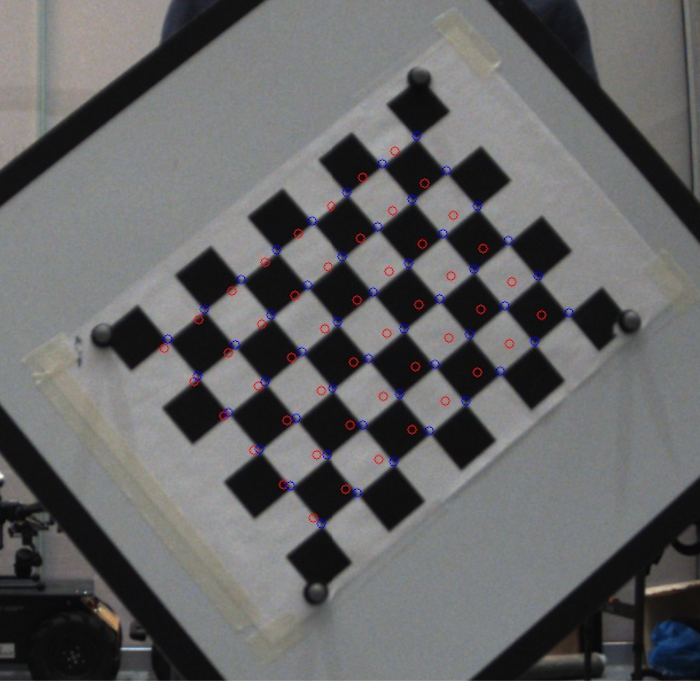}
  \end{subfigure}
  \begin{subfigure}[b]{0.16\textwidth}
    \includegraphics[trim={0cm 0cm 0cm 0cm},clip, width=\textwidth]{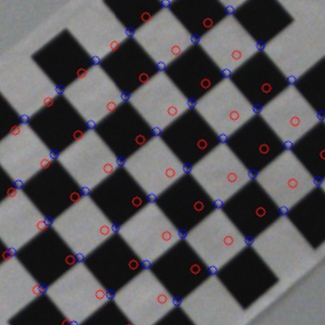}
  \end{subfigure}
  \begin{subfigure}[b]{0.16\textwidth}
    \includegraphics[trim={0cm 0cm 0cm 0cm},clip, width=\textwidth]{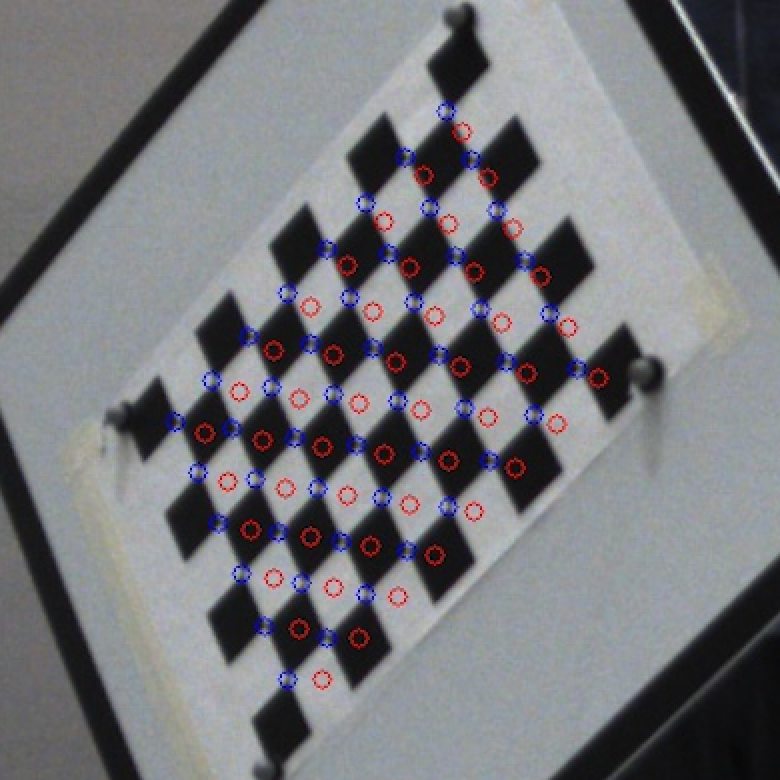}
  \end{subfigure}
  \begin{subfigure}[b]{0.16\textwidth}
    \includegraphics[trim={0cm 0cm 0cm 0cm},clip, width=\textwidth]{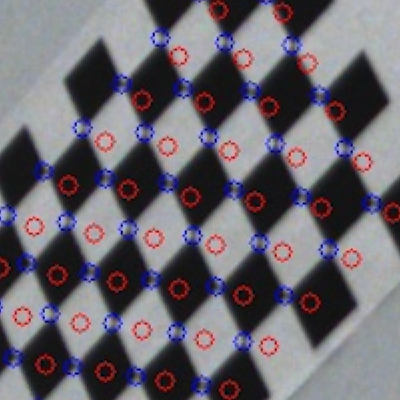}
  \end{subfigure}
\caption{Example experimental camera results. Red: initial estimates, blue: estimates after calibration}
\label{fig:results-real-cam}
\end{figure}

Camera calibration was also thoroughly validated with experimental data, where a similar reprojection error was observed for the experimental dataset relative to the simulation results. As shown in Table \ref{table:results-real-dia}, using 10 to 20 measurements of the diamond target, an average reprojection error of 0.64 pixels was achieved. Notice here that errors get larger with more measurements. First, it is worth pointing out that these are projection errors, not total calibration error as is shown in the simulation experiments, since there is no way to know the true calibration parameters. The reason this error term increases with more measurements is likely because the smaller experiments took the best images with the lowest target velocities and more direct views of the target. Fig. \ref{fig:results-real-cam} shows example camera calibration results. To compare results to the SOTA, we can convert our mean reprojection error to an approximate mean rotational error for each sensor calibration. Dividing the horizontal camera FOV by the image width, we get the degrees of rotation per pixel which can then be multiplied by the pixel error to get an approximate rotational error. Given an averaged reprojection error of 0.64 pixels, our estimated rotational error is computed at 0.04 degrees. This outperforms the targetless calibration methodology presented by Taylor et al.~\cite{Taylor2016} who achieved rotational errors between 0.1 degrees and 0.4 degrees on each of the three principle axes. Additionally, this method outperforms the edge-based calibration approach by Yuan et al.~\cite{yuan2021pixel}, achieving a residual error of approximately 1 pixel, despite the slight differences in our error measurement.

In summary, we proved experimentally that our approach outperforms SOTA targetless camera-lidar extrinsic calibration methods. We therefore provide the most accurate extrinsic calibration method that can be used to calibrate non-overlapping FOV lidar-camera systems. Being a target-based calibration method means guaranteed repeatable accuracy that other methods relying on the environment cannot provide. We also proved through simulation that our approach is robust to perturbation in the initial extrinsic estimates, and that our method can be used on various target designs, allowing researchers to experiment with different ways to extract measurements that better suits their data.

\subsection{Ablation Study}
\label{sec:ablation_study}
We evaluated the impact of the additional target alignment error term, \(T_E\) (Sec.~\ref{section:meth-math}), which we found to be crucial for aligning the estimated target frame with the actual target frame. As shown in Tab.~\ref{table:ablation-dia}, the additional error term significantly reduces the error and enhances optimization accuracy for both LiDAR and camera calibration.

\begin{table}[h]
\vspace{0.1cm}
\caption{Calibration errors of ablation study}
\label{table:ablation-dia}
\centering
\begin{tabular}{c c c c}
\hline
Lidar/Camera Test & No. Meas. & Error & Error \\
& & (w/o $T_E$) & (w/ $T_E$) \\
\hline
LiDAR Test 1 & 14 & 0.04753 (m) & 0.01679 (m) \\
LiDAR Test 2 & 23 & 0.03758 (m) & 0.0162282 (m) \\
Camera Test & 15 & 3.9042 (pixel) & 0.6932 (pixel) \\ 
\hline
\end{tabular}
\end{table}

\section{CONCLUSIONS}
\label{section:conclusion}

In this work, we show that a MCS can be used to perform extrinsic calibration of non-overlapping FOV lidar-camera systems without requiring a SLAM implementation. Our methodology results in a repeatable and accurate calibrations common to traditional target-based approaches, without the overlapping FOV constraints. We validate this with experimental and simulation data, while also validating that our implementation can be extended to various target shapes or feature extraction methods. This new extrinsic calibration methodology removes any limitations on sensor configuration while ensuring we can maintain precise data fusion to support any downstream data processing application in robotics. 

\label{section:references}
\bibliography{IEEE-conference-template-062824-CRV}
\bibliographystyle{ieeetr}

\end{document}